\pdfoutput=1

\documentclass[11pt]{article}

\usepackage[]{emnlp2022}

\usepackage{times}
\usepackage{latexsym}
\usepackage{hyperref}
\usepackage{tabularx}
\usepackage{booktabs}
\usepackage{multirow}
\usepackage{adjustbox}
\usepackage{caption}
\usepackage{subcaption}
\usepackage{tipa} 
\usepackage{enumitem}
\usepackage{amssymb}
\usepackage{amsthm}
\usepackage{pifont} 
\usepackage{array}
\usepackage{comment}
\usepackage{amsmath}
\usepackage{mathtools}
\usepackage{algorithm}
\usepackage{algpseudocode}
\newcommand{\cmark}{\ding{51}}
\newcommand{\xmark}{\ding{55}}

\newcommand{\bmark}{\ding{69}}
\newcommand{\emark}{\ding{72}}
\newcommand{\dmark}{\ding{63}}
\newcommand{\nmark}{\ding{248}}
\usepackage{xcolor}
\usepackage{hyperref}
\usepackage{xspace}

\usepackage[T4]{fontenc}
\usepackage{newunicodechar}
\newenvironment{tfour}{\fontencoding{T4}\selectfont}{}
\newcommand*{\yoruba}{Yor\`ub\'a\xspace}
\newunicodechar{ɔ}{\fon}
\newunicodechar{ɖ}{\fon}
\DeclarePairedDelimiter\ceil{\lceil}{\rceil}

\usepackage[T1]{fontenc}
\usepackage[utf8]{inputenc}
\usepackage{microtype}

\title{AfroLM: A Self-Active Learning-based Multilingual Pretrained Language Model for 23 African Languages}

\author{\normalsize Bonaventure F. P. Dossou$^{1,2,*}$, Atnafu Lambebo Tonja$^{3,*}$, Oreen Yousuf$^{4,*}$, Salomey Osei$^{5,*}$,  \\
\textbf{\normalsize Abigail Oppong$^{6,*}$, Iyanuoluwa Shode$^{7,*}$, Oluwabusayo Olufunke Awoyomi$^{8,*}$, Chris Chinenye Emezue$^{1,9,*}$} \\\\
\footnotesize
$^*$Masakhane NLP, $^1$Mila Quebec AI Institute, Canada, $^2$ McGill University, Canada, $^3$Instituto Politécnico Nacional, Mexico, \\ 
\footnotesize
$^4$Uppsala University, Sweden, $^5$Universidad de Deusto, Spain
$^6$Ashesi University, Ghana,
$^7$Montclair State University, USA,\\
\footnotesize
$^{8}$The College of Saint Rose, USA, $^{9}$Technical University of Munich, Germany
} 

\begin{document}

\maketitle
\begin{abstract}
In recent years, multilingual pre-trained language models have gained prominence due to their remarkable performance on numerous downstream Natural Language Processing tasks (NLP). However, pre-training these large multilingual language models requires a lot of training data, which is not available for African Languages. Active learning is a semi-supervised learning algorithm, in which a model consistently and dynamically learns to identify the most beneficial samples to train itself on, in order to achieve better optimization and performance on downstream tasks. Furthermore, active learning effectively and practically addresses real-world data scarcity. Despite all its benefits, active learning, in the context of NLP and especially multilingual language models pretraining, has received little consideration. In this paper, we present \textbf{AfroLM}, a multilingual language model pretrained from scratch on 23 African languages (the largest effort to date) using our novel self-active learning framework. Pretrained on a dataset significantly (14x) smaller than existing baselines, \textbf{AfroLM} outperforms many multilingual pretrained language models (AfriBERTa, XLMR-base, mBERT) on various NLP downstream tasks (NER, text classification, and sentiment analysis). Additional out-of-domain sentiment analysis experiments show that \textbf{AfroLM} is able to generalize well across various domains. We release the code source, and our datasets used in our framework at \url{https://github.com/bonaventuredossou/MLM_AL}.
\end{abstract}

\section{Introduction}
With the appearance of Transformer models \cite{transformers}, the field of Natural Language Processing (NLP) has seen the emergence of powerful multilingual pre-trained language models (MPLMs), such as mBERT \cite{devlin2018bert}, XLM-RoBERTa (XML-R) \cite{conneau2019unsupervised}, and mT5 \cite{mt5}. These prominent models have helped achieve state-of-the-art (SOTA) performance in many downstream NLP tasks such as named entity recognition (NER) \cite{alabi2022multilingual, masakhaner, devlin2018bert, conneau2019unsupervised}, text classification \cite{Ogueji-etal-small}, and sentiment analysis \cite{alabi2022multilingual, masakhaner, devlin2018bert, conneau2019unsupervised}. However they usually require a large amount of unlabeled text corpora for good performance: mBERT was trained on Wikipedia (2,500M words) and BookCorpus \cite{bookcorpus} (800M words) across 104 languages - 5 of which are African; mT5 supports 101 languages (13 African) and XLM-R supports 100 languages (8 African), and were trained on mC4 \cite{mt5} and CommonCrawl data \cite{commoncrawl}, respectively. This requirement for large-scale datasets contrasts sharply with the scarcity of available text corpora for African languages, which has pushed them into low-resource settings and largely excluded them from the pre-training phase of these large pre-trained models \cite{joshi, adelani2022few}. This exclusion, leads very often to a poor performance on languages unseen during pre-training \cite{alabi2022multilingual} which eventually leads to inability to carry out the required NLP task.
\begin{table*}[h!]
 \footnotesize
 \begin{center}
    \resizebox{\textwidth}{!}{
   \begin{tabular}{cccccccc}
   \toprule
\textbf{Languages}         & \textbf{Family}        &\textbf{Writing System}         & \textbf{African Region}  & \textbf{No of Speakers}    &\textbf{Initial \# of Sentences} &\textbf{Source}&\textbf{Size (MB)}\\\hline
Amharic (amh)         & Afro-Asiatic/Semitic & Ge'ez script & East            & 57M                & 655,079 &              \bmark,\dmark,\emark&279\\
Afan Oromo (orm)      & Afro-Asiatic/Cushitic        & Latin script  & East            & 37.4M                & 50,105 &               \dmark&9.87\\
Bambara (bam)    & NC/Manding           &	Latin, Arabic(Ajami), N'ko & West            & 14M                & 6,618 &               \bmark&~1.00\\
Ghomálá’ (bbj)   & NC/Grassfields       & Latin script & Central         & 1M                  & 4,841 &               \bmark&~0.50\\
Éwé (ewe)        & NC/Kwa               &	Latin (Ewe alphabet)  & West            & 7M                 & 5,615 &              \bmark&~0.50\\
Fon (fon)        & NC/Volta-Niger       & Latin script & West            & 1.7M                & 5,448 &              \bmark&~1.00\\
Hausa (hau)      & Afro-Asiatic/Chadic  &	
Latin (Boko alphabet) & West            & 63M                & 1,626,330 &               \bmark,\dmark,\emark&208\\
Igbo (ibo)       & NC/Volta-Niger       &	Latin (Önwu alphabet) & West            & 27M               & 437,737 &               \bmark,\dmark,\emark&63\\
Kinyarwanda (kin) & NC/Rwanda-Rundi       & Latin script & Central         & 9.8M               & 84,994 &               \nmark,\dmark,\bmark&37.70\\
Lingala (lin)     & NC/Bang                & Latin script & Central \& East & 45M               & 398,440 &               \bmark&45.90\\
Luganda (lug)    & NC/Bantu            & Latin script (Ganda alphabet)  & East            & 7M                 & 74,754&               \dmark,\bmark&8.34\\
Luo (luo)        & Nilo-Saharan           & Latin script & East            & 4M                    & 8,684 &               \dmark&1.29\\
Mooré (mos)      & NC/Gur              & Latin script  & West            & 8M                 & 27,908&               \bmark,\dmark&5.05\\
Chewa (nya)       & NC/Nyasa               & Latin script & South \& East   & 12M                & 8,000 &               \bmark&1.66\\
Naija (pcm)      & English-Creole       & Latin script  & West            & 75M                & 345,694 &              \bmark,\dmark,\emark&101\\
Shona (sna)       & NC/Bantu              & Latin script (Shona alphabet)  & Southeast       & 12M                & 187,810 &               \bmark,\dmark&32.80\\
Swahili (swa)    & NC / Bantu        &	Latin script (Roman Swahili alphabet)  & East \& Central & 98M               & 1,935,485&               \bmark,\dmark,\emark&276\\
Setswana (tsn)   & NC / Bantu            &	Latin (Tswana alphabet)  & South           & 14M                 & 13,958 &               \bmark,\dmark&2.21\\
Akan/Twi (twi)   & NC / Kwa              & Latin script  & West            & 9M                & 14,701&               \bmark&1.61\\
Wolof (wol)      & NC / Senegambia        & Latin (Wolof alphabet) & West            & 5M                  & 13,868&               \dmark&2.20\\
Xhosa (xho)       & NC/Zunda               &	Latin (Xhosa alphabet) & South           & 20M             & 93,288&               \bmark,\dmark&17.40\\
Yorůbá (yor)     & NC / Volta-Niger      &	Latin (Yorùbá alphabet)  & West            & 42M             & 290,999 &               \bmark,\dmark,\emark&45.9\\
isiZulu (zul)    & NC / Bantu            &	Latin (Zulu alphabet)  & South           & 27M                & 194,562&               \bmark,\dmark&33.70\\\hline

\end{tabular}
}
\caption{\textbf{Languages Corpora Details}. \textbf{Legends}: \cite{adelani2022few} $\rightarrow$ \bmark, \cite{alabi2022multilingual} $\rightarrow$ \dmark, \cite{Ogueji-etal-small} $\rightarrow$ \emark, \cite{kinnews} $\rightarrow$ \nmark.
}
\label{data_stats}
\end{center}
\end{table*}

Active learning is a semi-supervised machine learning algorithm that makes use of only a few initial training data points to achieve better performance of a given model \textbf{M}. The optimization is done by iteratively training \textbf{M}, and using another model \textbf{N}, usually referred to as the \textit{oracle}, to choose new training samples that will help \textbf{M} find better configurations while improving its performance (e.g., prediction accuracy). This makes active learning a prevalent paradigm to cope with data scarcity. The efficiency of active learning (i.e. its ability to produce better performance despite being trained on a smaller training data) has been proven in tasks such as biological sequence design~\cite{bio_gfn}, chemical sampling \cite{al_chem}, and Deep Bayesian (DB) approaches on image data \cite{deep_bayes_al}. Also, most of the work on deep active learning focuses on image classification with Convolutional Neural Networks (CNNs). It should be noted that active learning has been greatly explored and used to perform classification tasks, but not in language generation and understanding, and this is what we hope to address. 

A study of active learning in the context of NLP has been carried out by \cite{deep_bayes_al_1}. In their study, it is shown that active learning with DB networks coupled with uncertainty measures and acquisition function outperforms several i.i.d baselines. They showed that with only 20\% of samples labeled, their approach reached an accuracy of 98-99\% on the Named Entity Recognition (NER) task, while i.i.d tasks required 50\% of labelled data to achieve comparable performance. In their study on clinical texts, \cite{study_al} also proved that active learning algorithms outperformed other learning methods. \cite{ein-dor-etal-2020-active, tonneau-etal-2022-multilingual} on their works with BERT model(s) (for $n$ different languages, there were $n$ different BERT-based models) went further by showing that active learning works with a balanced and unbalanced dataset. They also showed that the different active learning methods performed relatively the same.

In our work, we fixed \textbf{M=N} (hence the title \textbf{self-active learning}). In our framework, we give \textbf{M} the ability to query itself, and use the knowledge acquired during each active learning round to construct new data points (from existing ones) that will be used for the next active learning round. 

We considered a diverse set of 23 African languages spread across the African continent. The selected languages are spoken in the south, central, east, and western regions of Africa. The languages cover four language families: Afro-Asiatic (e.g., Amharic, Hausa, Afan Omoro), Niger-Congo (NC) (e.g., Yorùbá, Bambara, Fon), English-Creole (Naija) and Nilo-Saharan (Luo) (see Appendix \ref{language_characteristics} for details). For each language, a dataset was collected from the news domain, which encompassed many topics such as health, politics, society, sport, environment, etc.

Our primary contribution to this work is our proposal of a \textbf{self-active learning framework} in which we pre-train the \textbf{biggest Multilingual African Language Model} (for the number of languages covered) to date, and we show that our setup is \textbf{very data-efficient and provides improvements} on downstream NLP tasks such as NER, text classification, and sentiment analysis (even on out-of-domain experiments).

\begin{figure*}[!ht]
    \centering
    \includegraphics[width=0.8\linewidth]{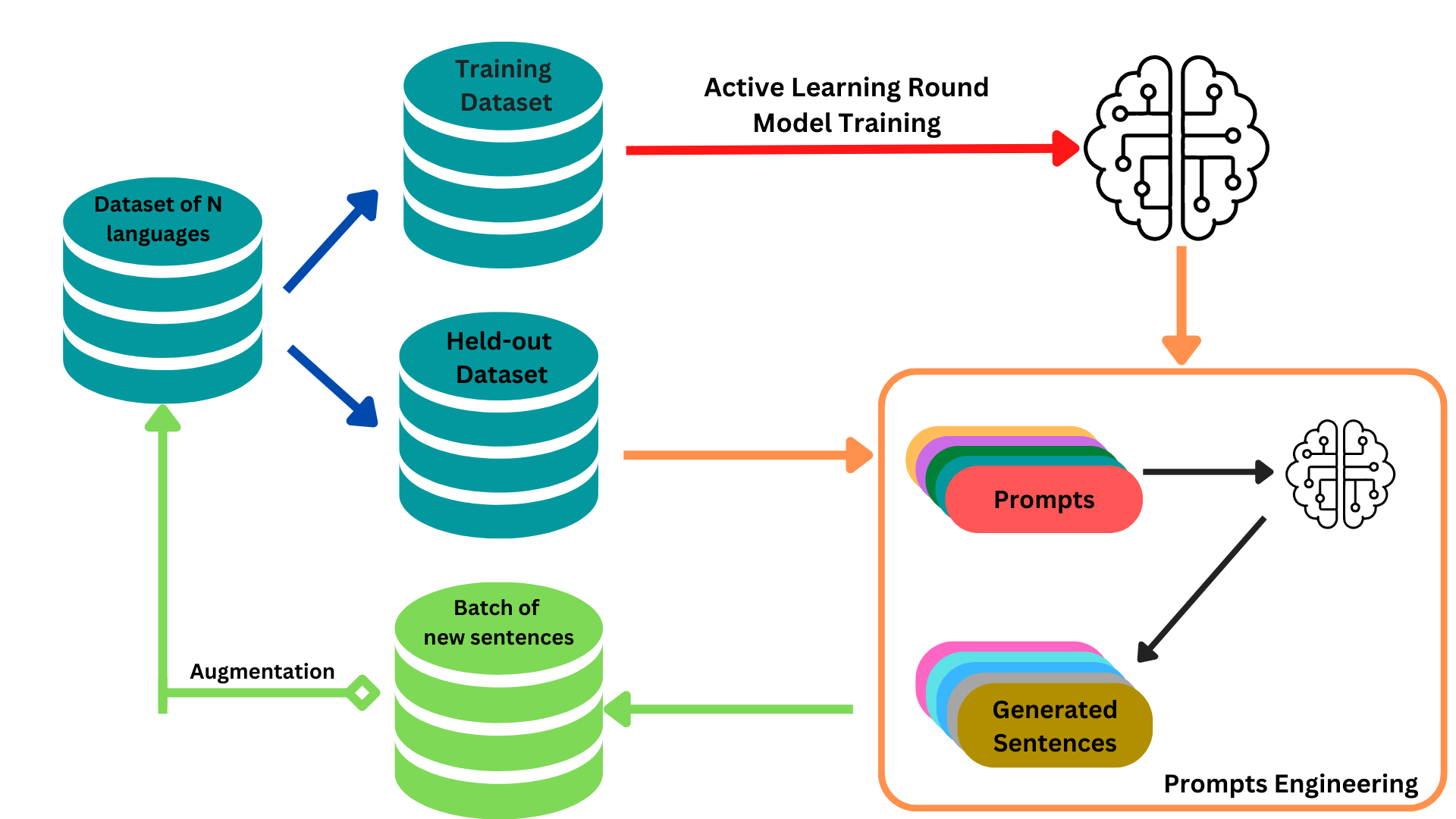}
    \caption{Self-Active Learning Framework). The process is designed in 4 stages (fully explained and detailed in Algorithm \ref{self_active}): (1) \textcolor{blue}{$\blacksquare$} Dataset split for current Active Learning round, (2) \textcolor{red}{$\blacksquare$} Active Learning round training, (3) \textcolor{orange}{$\blacksquare$} Generation of new sentence samples for the current round, and (4) \textcolor{green}{$\blacksquare$} Augmentation of the datasets of all languages.}
    \label{active_learning_framework}
\end{figure*}

\section{Related Works on MPLMs for African Languages}

\textit{Language adaptive fine-tuning (LAFT)} is one of the best approaches to adapt MPLMs to a new language. This entails fine-tuning an MPLM on monolingual texts of the said language with the same pre-training objective. However, this cannot be efficiently applied to African languages facing data-scarcity. \cite{Alabi-etal-2022-mert} proposed a new adaptation method called \textit{Multilingual adaptive fine-tuning (MAFT)}, as an approach to adapt MPLMs to many African languages with a single model. Their results show that MAFT is competitive to LAFT while providing a single model rather than many models that are specific for individual languages. Nevertheless, \citet{Alabi-etal-2022-mert}'s approach still works under the assumption that one does not need to train a model from scratch for languages in the low-resource settings, as they could benefit from high-resource languages. We find that this is not \textit{always} the case.

\cite{Ogueji-etal-small} introduced AfriBERTa, a multilingual language model trained on less than 1GB of data from 11 African languages. Training AfriBERTa from scratch showcased how African languages can benefit from being included in the pre-training stage of MPLMs. AfriBERTa produced competitive results compared to existing MPLMs (e.g., mBERT, XLM-R), and outperformed them on text classification and NER tasks. Rather than relying on high-resource languages for transfer-learning, AfriBERTa leverages the linguistic similarity between languages with low-resource settings to produce promising results. \cite{Ogueji-etal-small} empirically demonstrates that this is more beneficial to these languages and is crucial in assessing the viability of language models pre-trained on small datasets.
\raggedbottom

\cite{Nzeyimana-Rubungo-KinyaBERT} went beyond the linguistic taxonomy in creating KinyaBERT, a morphology-aware language model for Kinyarwanda. Trained on a 2.4GB corpus containing news articles from 370 websites registered between 2011 and 2021, KinyaBERT boasts a Transformer-like architecture that helps the representation of morphological compositionality. Their experiments outperformed solid baseline results for tasks such as NER and machine-translated GLUE on the Kinyarwanda language. These results demonstrated the effectiveness of not relying on transfer learning from high resource languages and rather explicitly incorporating morphological information of the African languages in their pre-training stage.

In the next section, we will describe our self-active learning framework, and the core details of our approach.
\section{Self-Active Learning Framework}
\label{framework_section}
In this section, we describe our self-active learning framework (Figure \ref{active_learning_framework}). In Algorithm \ref{self_active}, we present a single active learning loop. In our current work, our model is trained only with a Masked Language Modeling (MLM) objective \cite{conneau2019unsupervised, mlm, devlin2018bert}. We plan to further incorporate Translation Language Modeling (TLM) objective to improve translations of low-resource languages with relatively few thousands of data points \footnote{https://github.com/facebookresearch/XLM}. This will be useful for both supervised and unsupervised translation \cite{adelani2022few, conneau2019unsupervised}.

We used a shared Sentence Piece vocabulary with $250,000$ BPE codes. The subword shared vocabulary intends to improve alignment in the embedding space across languages (see languages description in Appendix \ref{language_characteristics} and corpora details in Table \ref{data_stats}) that are linguistically similar in features such as script/alphabet, morphology, etc. \cite{conneau2019unsupervised}, reflecting our focus languages. Additionally, \cite{conneau2019unsupervised} showed that scaling the size of the shared vocabulary (e.g. from 36,000 to 256,000) improved the performance of multilingual models on downstream tasks. Our vocabulary is defined jointly across all 23 languages and fixed during training, as opposed to random training and held-out dataset selection at each active learning round.

The motivations behind the randomness in the selection of the training and held-out datasets are: (1) to make efficient use of the limited dataset we have, and (2) to expose the model step by step, instead of simultaneously, to a variety of samples across different news sub-domains. We believe this would help in domain-shift adaptation and the robustness of the model. 

As extensively detailed in Algorithm \ref{self_active}, at each round we randomly select $m$ sentences per language, from the held-out dataset of the language. For a language, to generate a new sentence $s'$, given an original sentence $s$, we proceed as follows (more details can be found in Algorithm \ref{self_active}):
\begin{enumerate}
    \item select an initial ordered (left to right) set of words from $s$ as prompt,
    \item add a mask token at the end of the ordered set or sequence of words,
    \item query the model to predict the masked token,
    \item choose the best word, add it to the prompt,
    \item repeat 2-4 until we reach the length of $s$.
\end{enumerate}
The process described above will produce $m$ new data points that will be added to the language dataset. The new dataset obtained is used to re-train the model from scratch at the next active learning round.
\raggedbottom
\begin{algorithm}
\caption{Self-Active Learning Training Round}
\label{self_active}
\begin{algorithmic}
\Require
\State $\bullet$ Masked Language Modeling (MLM) objective $\pi_{\theta}$ with masking probability $p=0.15$
\State $\bullet$ Vocabulary $\mathcal{V}$, Model $\mathcal{M}$, Tokenizer $\mathcal{T}$
\State $\bullet$ Set of languages $\mathcal{L} = \bigcup_{i \in [1, 23]} \{l\}$
\State $\bullet$ Overall Dataset $\mathcal{D} = \bigcup_{l \in \mathcal{L}} \mathcal{D}_{l}$ with {$\mathcal{D}_{l}$} the dataset of language $l$
\State $\bullet$ Training Dataset $\mathcal{D}_{t}$ with $k\%$ randomly selected sentences from $\mathcal{D}_{l}, l \in \mathcal{L}$
\State $\bullet$  Held-out Dataset $\mathcal{H}$ with $1-k\%$ samples for each language: $\mathcal{H}$ = $\bigcup_{l \in L} \mathcal{H}_{l}$
\State $\bullet$ proportion $t$ of words to successively mask in a sentence (from left to right)
\Ensure
\State $\bullet$ Initialize $\mathcal{M}$, and $\mathcal{T}$ with $\mathcal{V}$
\State $\bullet$ $k \gets 80$
\State $\bullet$ $t \gets 15$
\State $\bullet$ Train $\mathcal{M}$ with policy $\pi_{\theta}$
\State Generate set $\mathcal{G}_{l}$ of new samples for each language:
\For{$l \in L$}
\State $\mathcal{G}_{l} \gets \{\}$
\State $\bullet$ Build $\mathcal{S}_{l}$ with $m$ = |$\mathcal{H}_{l}$| sentences randomly chosen from $\mathcal{H}_{l}$ \Comment{\textcolor{blue}{we choose $m$ this way to cope with small size datasets}}
\For{$s \in S_{l}$}
\State $n \gets len(s)$, $s = \bigcup_{i \in [1, n]} \{w_{i}\}$
\State $t_{s} \gets \ceil*{\frac{n*t}{100}} + 1$
\State $prompt \gets \bigcup_{i \in [1, n-t_{s}]} \{w_{i}\}$
\While{$t_{s} \neq 0$}
\State $prompt \gets prompt \cup \{$<mask>$\}$
\State $w_{p} \gets \mathcal{M}(prompt)$: \Comment{\textcolor{blue}{predicted masked word}}
\State $prompt \gets prompt \cup \{w_{p}\}$
\State $t_{s} \gets t_{s} - 1$
\EndWhile
\State $\mathcal{G}_{l} \gets \mathcal{G}_{l} \cup \{prompt\}$
\EndFor
\State $\mathcal{D}_{l} \gets \mathcal{D}_{l} \cup \mathcal{G}_{l}$
\Comment{\textcolor{blue}{new samples added to the language dataset}}
\EndFor
\end{algorithmic}
\end{algorithm}

\begin{table}[h!]
\footnotesize
 \begin{center}
\begin{tabular}{|c|c|c|}\hline
\textbf{Model} & \textbf{Hyper-parameters} & \textbf{Values} \\\hline
\multirow{12}{*}{\textbf{AfroLM-Large}} & sequence maximum length&256\\
& hidden size & 768\\
& attention heads& 6\\
& hidden layers & 10\\
& learning rate & 1e-4\\
& batch size & 32 \\
& \# of Parameters & 264M \\
& total initial training examples& 5,137,026 \\
& vocabulary size & 250,000 \\
& gradient accumulation steps & 8 \\
& warming steps & 40,000 \\
& training steps & 500,000 \\\hline
\end{tabular}
\end{center}
\caption{Hyper-parameters summary}
\label{hpm_table}
\end{table}

\begin{table*}[ht]
\footnotesize
 \begin{center}
   \begin{tabular}{ccccccc}
    \toprule
    \multirow{2}{*}{\textbf{Language}} & \textbf{In} & \textbf{In} & \textbf{In}&\textbf{In}&\textbf{In}& \\
    &\textbf{AfriBERTa?}&\textbf{AfroLM?}&\textbf{AfroXLMR}&\textbf{mBERT?}&\textbf{XLMR?}\\ 
    \midrule
    \texttt{amh} &\cmark & \cmark &\cmark &\xmark & \cmark\\
    \texttt{hau} &\cmark & \cmark &\cmark &\xmark &\cmark\\
    \texttt{ibo} &\cmark  & \cmark &\cmark &\xmark & \xmark\\
    \texttt{kin} &\cmark & \cmark &\cmark &\xmark &\xmark\\
    \texttt{lug} &\xmark & \cmark &\xmark &\xmark &\xmark\\
    \texttt{luo} &\xmark & \cmark &\xmark &\xmark &\xmark\\
    \texttt{pcm} &\cmark & \cmark &\cmark &\xmark &\xmark\\
    \texttt{swa} &\cmark & \cmark &\cmark &\cmark &\cmark\\
    \texttt{wol} &\cmark & \cmark &\cmark &\xmark &\xmark\\
    \texttt{yor} &\cmark & \cmark &\cmark &\cmark &\xmark\\
    \bottomrule
  \end{tabular}
    \caption{Information about languages included in each language model. We can notice that AfroLM includes the most of them.}
  \label{language_inclusion}
  \end{center}
\end{table*}

\begin{table*}[!ht]
\footnotesize
 \begin{center}
   \begin{tabular}{ccccccc}
    \toprule
    \multirow{2}{*}{\textbf{Language}} &  \multirow{2}{*}{\textbf{AfriBERTa-Large}}& \textbf{AfroLM-Large}& \textbf{AfroLM-Large}&\multirow{2}{*}{\textbf{AfroXLMR-base}}&\multirow{2}{*}{\textbf{mBERT}}&\multirow{2}{*}{\textbf{XLMR-base}} \\
    &&\textbf{(w/o AL)}&\textbf{(w/ AL)}\\ 
    \midrule
    \texttt{amh} &73.82 & 43.78& \textbf{73.84}&\textit{76.10}&00.00&70.96 \\
    \texttt{hau} & 90.17& 84.14& \textbf{91.09}&\textit{91.10}&87.34&87.44\\
    \texttt{ibo} &87.38& 80.24& \textbf{87.65}&\textit{87.40}&85.11 & 84.51\\
    \texttt{kin} & 73.78& 67.56& \textbf{72.84}&\textit{78.00}&70.98 & 73.93 \\
    \texttt{lug}& 78.85& 72.94 & \textbf{80.38}&\textit{82.90}&80.56&80.71\\
    \texttt{luo} & 70.23& 57.03 & \textbf{75.60}&\textit{75.10}&72.65&75.14\\
    \texttt{pcm} & 85.70& 73.23 & \textbf{87.05}&\textit{89.60}&87.78 & 87.39\\
    \texttt{swa} & 87.96& 74.89 & \textbf{87.67}&\textit{88.60}&86.37 & 87.55\\
    \texttt{wol} & 61.81& 53.58 & \textbf{65.80} & \textit{67.40}&66.10 & 64.38\\
    \texttt{yor} & 81.32& 73.23 & \textbf{79.37}&\textit{82.10}&78.64 & 77.58\\
    \midrule
    avg &79.10& 68.06& \textbf{80.13}& \textit{81.90}& 71.55& 79.16\\
    avg (excl. \texttt{amh}) &79.69& 70.76& \textbf{80.83}& \textit{82.54}& 79.50& 80.07\\
    \bottomrule
  \end{tabular}
    \caption{\textbf{NER Performances:} F1-scores on languages test sets after 50 epochs averaged over 5 seeds. These results cover all 4 tags in the MasakhaNER dataset: \textbf{PER}, \textbf{ORG}, \textbf{LOC}, \textbf{DATE}. XLM-R and mBERT results
obtained from \cite{adelani2021masakhaner}. \textbf{AfroLM-Large (w/ AL)} outperforms AfriBERTa, and the initial MasakhaNER baselines. \textbf{The bold numbers represent the performance of the model with the lowest pretrained data}. AfroXMLR-base = XLMR-Large + MAFT \cite{alabi2022multilingual} with 272M parameters. MAFT gives similar performance to individual LAFT models \cite{alabi2022multilingual} (LAFT results in single model per language).}
  \label{ner_results}
  \end{center}
\end{table*}

\begin{table*}[t]
\begin{center}
\footnotesize
\resizebox{\textwidth}{!}{%
\begin{tabular}{lrrrrrrrrrrr|c}
\toprule
\textbf{Model} & \textbf{bam} & \textbf{bbj} & \textbf{ewe} & \textbf{fon} & \textbf{mos} & \textbf{nya} & \textbf{sna} & \textbf{tsn} & \textbf{twi}& \textbf{xho} & \textbf{zul} & \textbf{AVG}  \\
\midrule
\multicolumn{7}{l}{MPLMs pre-trained on from scratch on African Languages} \\
AfriBERTa-Large & 78.60 & 71.00 & 86.90 & 79.90 & 71.40 & 88.60 & 92.40 & 83.20 & 75.70 & 85.00 & 81.70 & $81.31$ \\
\textbf{AfroLM-Large (w/ AL)} & \textbf{80.40} & \textbf{72.91} & \textbf{88.14} & \textbf{80.48} & \textbf{72.14} & \textbf{90.25} & \textbf{94.46} & \textbf{85.38} & \textbf{77.89} & \textbf{87.50} & \textbf{86.31} & \textbf{83.26} \\
\bottomrule
\multicolumn{7}{l}{MPLMs adapted to African Languages} \\
\textit{AfroXLMR-base} & \textit{79.60} & \textit{73.30} & \textit{89.20} & \textit{82.30} & \textit{74.40} & \textit{91.90} & \textit{95.70} & \textit{87.70} & \textit{78.90} & \textit{88.60} & \textit{88.40} & \textit{84.55} \\
mBERT & 78.90 & 60.60 & 86.90 & 79.90 & 71.40 & 88.60 & 92.40 & 86.40 & 75.70 & 85.00 & 81.70 & 80.68 \\
XLMR-base & 78.70 & 72.30 & 88.50 & 81.90 & 72.70 & 89.90 & 93.60 & 86.10 & 78.70 & 87.00 & 84.60 & 83.09 \\
\bottomrule
    \end{tabular}
    }
\caption{\textbf{NER Baselines on MasakhaNER2.0 \cite{masakhaner2}}. We compare MPLMs trained from scratch on African languages, and MPLMs adapted to African Languages. The average of scores are over 5 runs. The bold numbers represent the performance of the model with the lowest pretrained data.}
\label{masakhaner2_baselines}
  \end{center}
\end{table*}

\begin{table*}[ht]
\footnotesize
 \begin{center}
   \begin{tabular}{ccccccc}
    \toprule
    \multirow{2}{*}{\textbf{Language}} & \multirow{2}{*}{\textbf{AfriBERTa-Large}}& \textbf{AfroLM-Large}&\textbf{AfroLM-Large}\\
    &&\textbf{(w/o AL)}&\textbf{(w/ AL)} \\ 
    \midrule
    \texttt{hau} & 90.86 & 85.57&\textbf{91.00}\\
    \texttt{yor} & \textit{83.22} & 75.30& \textbf{82.90}\\
    \bottomrule
  \end{tabular}
    \caption{\textbf{Text Classification Performances:} F1-scores on the languages test sets. \textbf{The bold numbers represent the performance of the model with the lowest pretrained data}.}
  \label{text_class_results}
  \end{center}
\end{table*}

\begin{table*}[h!]
\footnotesize
 \begin{center}
   \begin{tabular}{cc}
    \toprule
    \textbf{Models} & \textbf{Yoruba F1-score}\\ 
    \midrule
    \textbf{AfroLM-Large (w/o AL)} &\\
    \texttt{Movies} & 83.12\\
    \texttt{Twitter} $\rightarrow$ \texttt{Movies} & 41.28 \\
    \midrule
    \textbf{AfroLM-Large (w/ AL)} &\\
    \texttt{Movies} & \textbf{85.40}\\
    \texttt{Twitter} $\rightarrow$ \texttt{Movies} & \textbf{68.70} \\
    \midrule
    \textbf{AfriBERTa-Large} & \\
    \texttt{Movies} & 82.70\\
    \texttt{Twitter} $\rightarrow$ \texttt{Movies} & 65.90 \\
\bottomrule
  \end{tabular}
    \caption{\textbf{Out-Of-Domain Sentiment Analysis Performance:} F1-scores on YOSM test set after 20 epochs averaged over 5 seeds. \textbf{The bold numbers represent the performance of the model with the lowest pretrained data}.}
  \label{sentiment_results}
  \end{center}
\end{table*}

\section{Experiments, Results and Discussion}
\paragraph{Experiments:}We use the XLM-RoBERTa (XLM-R) architecture in our experiments based on previous works utilizing the model to achieve state-of-the-art performance in various downstream tasks. Following the work and results of \cite{Ogueji-etal-small}, we trained XLM-R-based models from scratch. In our current work we trained the model with $3$ self-active learning rounds (we stopped at 3 due to computational resources). We used 80\% and 20\% of languages data for the training and held-out datasets respectively. We designed 2 versions of AfroLM: \textit{AfroLM-Large (without self-active learning)} and \textit{AfroLM-Large (with self-active learning)} with the hyper-parameters specified in Table \ref{hpm_table}. All training experiments were done using the HuggingFace Transformers library \cite{huggingface}.

\textbf{AfroLM (without self-active learning)} is one of our baselines. We trained an XLM-R model on the entire dataset, and the held-out dataset was just used for evaluation. For \textbf{AfroLM-Large} models, we used Google Cloud with a single 48GB NVIDIA A100 GPU. An active learning round took $\approx$ 260 hours of training. We evaluated \textbf{AfroLM-Large} models on three downstream tasks:
\begin{itemize}
    \item \textbf{NER}: we evaluated the performance of our model pre-trained using our self-active learning framework on the MasakhaNER dataset \cite{masakhaner}. The dataset contains ten African languages: Amharic, Hausa, Igbo, Kinyarwanda, Luganda, Luo, Nigerian Pidgin, Swahili, Wolof, and Yorùbá. \cite{masakhaner, alabi2022multilingual} also provided strong baselines with pre-trained language models like mBERT and XLM-R on MasakhaNER.
    \item \textbf{Text Classification}: we tested our models on Hausa and Yorùbá news text classification dataset from \cite{hedderich-etal-2020-transfer}, where the authors have also built strong baselines on mBERT and XLM-R models.
    \item \textbf{Sentiment Analysis}: we tested the the out-of-domain performance of our model in two domains different from news:
    \begin{enumerate}
        \item \texttt{Movies}: we directly fine-tuned and evaluated \textbf{AfroLM-Large} on the YOSM dataset \cite{yosm}, which contains reviews of Yorùbá movies.
        \item \texttt{Twitter $\rightarrow$ Movies}: in this setup, we finetuned on the training and validation set of NaijaSenti \cite{naija_senti}, and evaluated on YOSM. NaijaSenti contains human annotated tweets in Hausa, Yoruba, Igbo and Nigerian Pidgin. However, we were not able to evaluate \textbf{AfroLM-Large} on it because the authors have not yet released the test set.
    \end{enumerate}
\end{itemize}

\paragraph{Results \& Discussion:}Tables \ref{data_stats} and \ref{language_inclusion} show that our framework includes a large variety of African Languages. Table \ref{ner_results}, and Table \ref{masakhaner2_baselines} (with 11 additional languages from MasakhaNER 2.0 dataset \cite{masakhaner2}) show the results of our method in comparison with other baselines on NER task. We can notice that \textbf{AfroLM-Large (w/ AL)} outperforms AfriBERTa-Large, mBERT and XLMR-base ($\approx$ 2.5 TB of data); while being pre-trained on significantly smaller dataset ($\approx$ 0.73 GB (80\% of 0.91 GB initial dataset)). AfriBERTa-Large has been pretrained from scratch on 11 African languages, while mBERT and XLMR-base (with existing pretrained weights) were finetuned on the MasakhaNER dataset. 

Table \ref{text_class_results} and Table \ref{sentiment_results} show that, on the text classification and sentiment analysis tasks, our method outperforms many existing baselines. Additionally, out-of-domain experiments and analyses show that our method is robust and provides good results in out-of-domain settings.

While AfroXLMR-base in average, slightly outperforms our approach, it is important to notice that it has been pretrained on a dataset 14x bigger than our set. Furthermore, \textbf{AfroLM-Large} has been trained on $\approx$ 0.73 GB of data (80\% of 0.91 GB initial dataset), which is less than the size of the corpus used to train AfriBERTa (0.939 GB). This allows us to confidently affirm that our approach is data-efficient, while being very competitive.
\raggedbottom

It is important to note that the margin of performance from \textbf{AfroLM-Large (w/ AL)} does not come from the fact that it has been trained on more languages. Our results show that \textbf{AfroLM-Large (w/ AL)} outperforms models trained on significantly larger datasets and number of languages. Moreover, the comparison of \textbf{AfroLM-Large (w/ AL)} to \textbf{AfroLM-Large (w/o AL)} shows a significant improvement in performance, which implies that our self-active learning framework is efficient, and leads to a better performance. This is expected, because the idea of our self-active learning (and of active learning in general) is that \textbf{AfroLM} consistently and dynamically, identifies during the training phase, the most beneficial sample(s) to learn from in order to boost the performance.

In our current algorithm, a sentence sample is generated by \textit{\textbf{iterative next-token prediction}}: the generated sentence is the result of the concatenation of each best token. Diversity in sample generation and selection is paramount, and we believe, could improve the performance of our framework. In the limitation section (section \ref{limitation}), we propose a way of selecting \textit{diverse} sentences (after sentence generation). We also propose a new weighted loss, that we believe will be more balanced across the entire dataset.

\section{Future works and Conclusion}
In conclusion, we propose \textbf{AfroLM}, a self-active learning-based multilingual language model supporting 23 African Languages; the largest to date. Our language datasets are collected from the news domain and span across different parts of the African continent. Our experimental results on NLP downstream tasks (NER, text classification, and out-of-domain sentiment analysis), prove the data-efficiency of \textbf{AfroLM} (as it has been trained on a dataset 14x smaller than its competitors), and its competitiveness as it outperforms many MPLMs (AfriBERTa, mBERT, XLMR-base) while being very competitive to \textit{AfroXLMR-base}. We also show that \textbf{AfroLM} is also able to generalize across various domains. For future work, we intend to: (1) explore and understand the relationship between the number of active learning steps and the MPLMs performance on downstream tasks, and (2) integrate a new weighted loss, and more diversity in new data points generation and selection as we explain in the limitation section (see section \ref{limitation}). Our datasets, and source code are publicly available at \url{https://github.com/bonaventuredossou/MLM_AL}.
\section{Limitations and Approach of Solution}
\label{limitation}
Currently, the loss of the model across the training dataset (across all 23 languages), appears to be the average of the individual (cross-entropy) losses. Due to the disparate sizes of our corpora per language, the training will be biased toward the languages whose sizes predominate the training set. Therefore, we suggest a strategy to re-weight the cross entropy loss per language by the ratio of the size of the dataset for that language to the size of the entire training set:


\[\mathcal{L} = \frac{1}{N} \sum_{l} |\frac{\mathcal{D}_{l}}{\mathcal{D}}| \mathcal{L}_{l}\]
where $|\frac{\mathcal{D}_{l}}{\mathcal{D}}|$ is the weight of the training dataset of the language $l$, $\mathcal{L}_{l}$ is the loss of the model on a given language $l$, and $N$ is the total number of languages (23 in our case). We believe this adjusts well overall loss by using the right weighted loss of each language, which can be seen as their respective contribution to the general loss.

Another limitation of our current framework is that the samples that are generated from prompts might not be diverse. Given a batch $\mathcal{B}$ of generated samples, and a set $\mathcal{S}$ of initial samples, we want the samples selected to be substantially different from the majority of samples present in $\mathcal{S}$.
We think that performing the following two steps will help to ensure this:
\begin{enumerate}
    \item Increasing the number of words, in a sentence, to be masked: this implies that the length of the prompt is shortened, and that we provide less (or short) context in the input to our model. Long-range semantics is still a challenge in natural language generation and understanding, and large language models (GPT-2, DialoGPT) have insufficiently learned the effect of distant words on next-token prediction \cite{malkin-etal-2022-coherence}. Therefore, we believe that providing a short context will increase the choices of the model and lead to the generation of more various tokens. This has been shown by \cite{malkin-etal-2022-coherence} where they also introduced the \textit{coherence boosting} approach to increase the focus of a language model on a long context.
    \item Using the Word Error Rate (WER) as a simple diversity measurement. The WER is an adaptation of the Levenshtein distance (also called edit distance), working at the word level instead of the phoneme level. Ideally, we want high WER. Let $W = \bigcup_{i \in [1, t_{s}]} \{w_{i}\}$, the set of words from a sentence $s$ that we cut off for the next-token prediction loop described in section \ref{framework_section} and in Algorithm \ref{self_active}. Let $W' = \bigcup_{i \in [1, t_{s}]} \{w'_{i}\}$, the set of words predicted by the model. Then, for a pair $(s, s')$ of the original sentence and new generated sentence ($s' = prompt \cup W'$), we can define a diversity score $d_{s, s'} = WER(W, W')$. Given the definition of $d$, for a language $l$, we can define a diverse batch \[B^{l}_{diverse} = \bigcup_{i \in [1, |\mathcal{H}_{l}|]} \{s'_{i} \hspace{1mm} |\hspace{1mm} d_{s_{i}, s'_{i}} \geq t \}\] where $t$ is an hyper-parameter, representing an error threshold. $t$ can be tuned because a small $t$ will result in a less diverse batch, while a very huge value will result in an empty or almost empty batch.
\end{enumerate}

\section{Ethics Statement}
As any modern technology, machine learning algorithms are subject to potential dual good or bad usage. Our work is motivated by the desire of making AI (in general, NLP in particular) applications to be inclusive to the low-resourced languages (which are the vast majority of existing living languages), hence benefiting to humanity and society. We strongly discourage bad and unethical use of our work (and its derivations).

\bibliography{anthology,custom}
\bibliographystyle{acl_natbib}

\appendix
\section{Language Characteristics}
\label{language_characteristics}
\paragraph{Amharic (amh)} also called Amarinya or Amerigna, is a Semitic language, an official language of Ethiopia, and is also spoken in Eritrea. Amharic is written with a modified version of the Ge'ez script, known as Fidel, consisting of 33 basic characters, each of them with at least 7 vowel sequences. Unlike Central and Northwest Semitic languages such as Arabic, Hebrew and Assyrian Aramaic, Amharic is written from left to right. The language has a variety of local dialects, all of which are mutually intelligible. There are three major dialects: Gondar, Gojjami, and Showa. There are specially marked differences in pronunciation, vocabulary, and grammar between the northern Gojjami and the southern Showa dialects. 

\paragraph{Afan Oromo (oro)} is an Afroasiatic language that belongs to the Cushitic branch spoken by about 30 million people in Ethiopia, Kenya, Somalia and Egypt, and it is the third largest language in Africa. The Oromo people are the largest ethnic group in Ethiopia and account for more than 40\% of the population. They can be found all over Ethiopia, and particularly in Wollega, Shoa, Illubabour, Jimma, Arsi, Bale, Hararghe, Wollo, Borana and the southwestern part of Gojjam\footnote{https://omniglot.com/writing/oromo.htm}. Afan Oromo is written with a Latin alphabet called Qubee. Like most other Ethiopian languages, whether Semitic, Cushitic, or Omotic, Oromo has a set of ejective consonants, that is, voiceless stops or affricates that are accompanied by glottalization and an explosive burst of air. Afan Oromo has another glottalized phone that is more unusual, an implosive retroflex stop, "dh" in Oromo orthography, a sound that is like an English "d" produced with the tongue curled back slightly and with the air drawn in so that a glottal stop is heard before the following vowel begins. It is retroflex in most dialects, though it is not strongly implosive and may reduce to a flap between vowels\footnote{\url{https://en.wikipedia.org/wiki/Oromo\_language}}. In the Qubee alphabet, letters include the digraphs ch, dh, ny, ph, sh. Gemination is not obligatorily marked for digraphs, though some writers indicate it by doubling the first element: qopphaa'uu 'be prepared'. Afan Oromo has five vowel phonemes, i.e., sounds that can differentiate word meaning. They can be short or long. The length of the vowel makes a difference in word meaning e.g., laga ‘river’ and laagaa ‘roof of the mouth’. Afan Oromo has 25 consonant phonemes, i.e., sounds that make a difference in word meaning. Like its close relative, Somali, native Oromo words do not have the consonants /p/, /v/, and /z/.

\paragraph{Bambara (bam)} is a Western Mande language with about 14 million speakers mainly in Mali, and also in Senegal, Niger, Mauritania, Gambia and Côte d'Ivoire. It is spoken principally among the Bambara ethnic group in Mali, where it is the national language and the most widely understood one. Bambara is usually written with the Latin alphabet, though the N'Ko and Arabic alphabets are also used to some extent. It uses seven vowels a, e, \textepsilon, i, o, \textopeno,  and u each of which can be nasalized, pharyngealized and murmured, giving a total number of 21 vowels.

\paragraph{Ghomalá’ (bbj)} is a major Bamileke language spoken in Cameroon. It is spoken by an estimated 1.1 million people in two main population groups. 

\paragraph{Éwé (ewe)} is a language spoken in Togo and southeastern Ghana by approximately 20 million people mainly in West Africa in the countries of Ghana, Togo, and Benin. It is recognised as a national language in Ghana, where English is the official language, and in Togo, where French is the official language. 'Ewe' is also the name of the tribal group that speaks this language. Éwé has three distinguishable dialects. Most of the differences among the dialects have to do with phonology. All dialects are mutually intelligible. Éwé is written in the African reference alphabet, first proposed by a UNESCO-organized conference in 1978. It is a version of the Latin alphabet adapted to represent Éwé sounds. Some sounds are represented by two-letter sequences, e.g., dz, ts, gb, kp, ny. Éwé has seven oral and five nasal vowels. Nasal vowels are produced by lowering the soft palate so that air escapes both through the mouth and the nose. Nasal vowels are marked by a tilde.

\paragraph{Fon (fon)} also known as Fongbé is a native language of Benin Republic. It is spoken in average by 1.7 million people. Fon belongs to the \textit{Niger-Congo-Gbe} languages family. It is a tonal, isolating and left-behind language according to \cite{joshi}, with an \textit{Subject-Verb-Object} (SVO) word order. Fon has about 53 different dialects, spoken throughout Benin \cite{grammaire, fon_phonology, ethnologue}. Its alphabet is based on the Latin alphabet, with the addition of the letters: \begin{tfour}\m{o}\end{tfour}, \begin{tfour}\m{d}\end{tfour}, \begin{tfour}\m{e}\end{tfour}, and the digraphs gb, hw, kp, ny, and xw. There are 10 vowels phonemes in Fon: 6 said to be closed [i, u, ĩ, ũ], and 4 said to be opened [\begin{tfour}\m{e}\end{tfour}, \begin{tfour}\m{o}\end{tfour}, a, ã]. There are 22 consonants (m, b, n, \begin{tfour}\m{d}\end{tfour}, p, t, d, c, j, k, g, kp, gb, f, v, s, z, x, h, xw, hw, w). Fon has two phonemic tones: high and low. High is realized as rising \textit{(low–high)} after a consonant. Basic disyllabic words have all four possibilities: \textit{high-high}, \textit{high-low}, \textit{low-high}, and \textit{low-low}. In longer phonological words, like verb and noun phrases, a high tone tends to persist until the final syllable. If that syllable has a phonemic low tone, it becomes falling \textit{(high–low)}. Low tones disappear between high tones, but their effect remains as a downstep. Rising tones \textit{(low–high)} simplify to high after high (without triggering downstep) and to low before high \cite{grammaire, fon_phonology}.

\paragraph{Hausa (hau)} belongs to the West Chadic branch of the Afro-Asiatic language family. It is one of the largest languages on the African continent, spoken as a first language by the original Hausa people and by people of Fula ancestry. Hausa is the majority language of much of northern Nigeria and the neighboring Republic of Niger. In addition, there is a sizable Hausa-speaking community in Sudan\footnote{https://www.mustgo.com/worldlanguages/hausa/}.
It has an alphabet of 29 letters containing 5 vowels and 24 consonants. Hausa alphabet is a Latin script/Roman alphabet/English letters except (x, v, p, and q) and also added six extra letters (\texthtb, \texthtd, \texthtk, sh, ts and \begin{tfour}\m{y}\end{tfour} \cite{adelani2021masakhaner}. Hausa is an agglutinative language, i.e., it adds suffixes to roots for expressing grammatical relations without fusing them into one unit, as is the case in Indo-European languages.

\paragraph{Ìgbò (ibo)} is one of the largest languages of West Africa, is spoken by 18 million people in Nigeria. It belongs to the Benue-Congo group of the Niger-Congo language family. The language is thought to have originated around the 9th century AD in the area near the confluence of the Niger and Benue rivers, and then spread over a wide area of southeastern Nigeria \footnote{https://www.mustgo.com/worldlanguages/igbo/}. Igbo is a national language of Nigeria and is also recognised in Equatorial Guinea. Igbo is written in an expanded version of the Latin alphabet. Igbo is made up of many different dialects which aren't mutually intelligible to other Igbo speakers at times. 

\paragraph{Kinyarwanda (kin)} is part of the Bantu sub-group of the central branch
of the Niger-Congo language family. It is closely related to Kirundi, the language of Burundi. The Rwanda language is mutually intelligible with Kirundi, which is spoken in neighboring Burundi\footnote{https://nalrc.indiana.edu/doc/brochures/kinyarwanda.pdf}. It has only 18/19 consonants, as X and Q are not found in the alphabet. L is often replaced by R, but due to the appearance of imported words in the language, that is not always the case. It has five vowel phonemes, i.e., sounds that make a difference in word meaning.

\paragraph{Lingala (lin)} is a Central Bantu language that belongs to the
largest African languages phylum: the Niger-Congo. Lingala is spoken as a first, second, and third language primarily in the Democratic Republic of Congo (DRC), the Republic of Congo (Congo-Brazzaville), and in parts of five neighboring central African states: Northwestern Angola, eastern Gabon, southern Central African Republic, and southwestern Sudan. The estimated number of speakers ranges from twenty to twenty five million\footnote{https://nalrc.indiana.edu/doc/brochures/lingala.pdf}. It is written with the Latin alphabet. The seven vowels are represented by five symbols. The orthographic symbols 'e' and 'o' each represent two sounds. There are two tones in Lingala. High tone is represented with an acute accent, while low tone is unmarked.

\paragraph{Luganda (lug)} is a Bantu language spoken in the African Great Lakes region. It is one of the major languages in Uganda and is spoken by more than 10 million Baganda and other people principally in central Uganda including the capital Kampala of Uganda. Its alphabet is composed of twenty-four letters; 18 consonants (b, p, v, f, m, d, t, l, r, n, z, s, j, c, g, k, ny, \textipa{\ng}), 5 vowels ( a, e, i, o, u) and 2 semi-vowels(w, y). Since the last consonant \textipa{\ng}) does not appear on standard typewriters or computer keyboards, it is often replaced by the combination ng'. All consonants are pronounced as if with letter ‘a’ or ‘ah’ at the end. For example, bah, cah, jah, gah, kah, mah, pah, lah, zah, e.t.c

\paragraph{Luo (luo)} are spoken by the Luo peoples in an area ranging from southern Sudan to southern Kenya, with Dholuo extending into northern Tanzania and Alur into the Democratic Republic of the Congo. Luo has a CVC or VC structure—consonant/vowel/consonant or vowel/consonant. This is unlike Bantu languages, where words must end in a vowel. Luo language is therefore more similar to English articulation, while Bantu languages are more like Italian\footnote{https://owlcation.com/humanities/Luo-language-of-Kenya-Conversation-Basics}. 

\paragraph{Mooré (mos)} is a Gur language of the Oti–Volta branch and one of two official regional languages of Burkina Faso. It is the language of the Mossi people, spoken by approximately 8 million people in Burkina Faso, plus another 1M+ in surrounding countries such as Ghana, Cote D'ivoire, Niger, Mali and Togo as a native language, but with many more L2 speakers. Mooré is spoken as a first or second language by over 50\% of the Burkinabè population.

\paragraph{Chewa (nya)} is a Bantu language spoken in much of Southern, Southeast and East Africa, namely the countries of Malawi and Zambia, where it is an official language, and Mozambique and Zimbabwe where it is a recognised minority language. Chewa has five vowel sounds: /a, \textepsilon, i, \textopeno, u/; these are written a, e, i, o, u. 

\paragraph{Naija (pcm)} is an English-based creole language spoken as a lingua franca across Nigeria. The language is sometimes referred to as "Pijin" or Broken (pronounced "Brokun").

\paragraph{Shona (sna)} is a Bantu language of the Shona people of Zimbabwe. All syllables in Shona end in a vowel. Consonants belong to the next syllable. For example, mangwanani ("morning") is syllabified as ma.ngwa.na.ni; "Zimbabwe" is zi.mba.bwe. No silent letters are used in Shona.

\paragraph{Swahili (swa)} also known by its native name Kiswahili, is a Bantu language and the native language of the Swahili people native primarily to Tanzania. Swahili has become a second language spoken by tens of millions in four African Great Lakes countries (Kenya, DRC, Uganda, and Tanzania), where it is an official or national language, while being the first language for many people in Tanzania especially in the coastal regions of Tanga, Pwani, Dar es Salaam, Mtwara and Lindi. Standard Swahili has five vowel phonemes: /a/, /\textepsilon/, /i/, /\textopeno/, and /u/. 

\paragraph{Setswana (tsn)} is a Bantu language spoken in Southern Africa by about 14 million people. Setswana is an official language and lingua franca of Botswana and South Africa. 

\paragraph{Akan/Twi} is a dialect of the Akan language spoken in southern and central Ghana by several million people, mainly of the Akan people, the largest of the seventeen major ethnic groups in Ghana. Twi excludes consonants such as c, j, q, v, x and z. It has 15 consonants and 7 vowels. Apart from [a], [e], [i], [o] and [u], Twi also has 2 additional vowels; [\textepsilon] and [\textopeno].

\paragraph{Wolof (wol)} is a language of Senegal, Mauritania, and the Gambia, and the native language of the Wolof people. Wolof is the most widely spoken language in Senegal, spoken natively by the Wolof people (40\% of the population) but also by most other Senegalese as a second language.

\paragraph{Xhosa (xho)} also isiXhosa as an endonym, is a Nguni language and one of the official languages of South Africa and Zimbabwe. The Xhosa language employs 26 letters from the Latin alphabet. Xhosa has an inventory of ten vowels: [a], [\begin{tfour}\m{e}\end{tfour}~e], [i], [\begin{tfour}\m{o}\end{tfour}~o] and [u] written a, e, i, o and u in order, all occurring in both long and short. The /i/ vowel will be long in the penultimate syllable and short in the last syllable.
\paragraph{\yoruba (yor)} has 25 Latin letters without the Latin characters (c, q, v, x and z) and with additional letters ({\d e}, gb,{\d s}, {\d o}).\yoruba is a tonal language with three tones: low ("\textbackslash"), middle ("\textemdash", optional) and high ("\slash"). The Latin letters ⟨c⟩, ⟨q⟩, ⟨v⟩, ⟨x⟩, ⟨z⟩ are not used as part of the official orthography of Standard \yoruba, however, they exist in several \yoruba dialects. The tonal marks and underdots are referred to as diacritics and they are needed for the correct pronunciation of a word. \yoruba is a highly isolating language and the sentence structure follows subject-verb-object \cite{adelani2021masakhaner}.

\paragraph{Zulu (zul)} is the mother tongue of the Zulu people, South's Africa largest ethnic group, who created an empire in the 19th century.Zulu has a 7-vowel system. Each vowel can be long or short. Zulu has close to 50 consonants including clicks, ejectives and implosives. Clicks originated in Khoisan languages and then spread into some neighboring Bantu ones. In Zulu they have three places of articulation: central alveolar, lateral alveolar and palatal combined with five accompaniments (plain, aspirated, voiced, nasal, and voiced nasal).

\end{document}